\documentclass[10pt,twocolumn,letterpaper]{article}

\usepackage{cvpr}
\usepackage{times}
\usepackage{epsfig}
\usepackage{graphicx}
\usepackage{amsmath}
\usepackage{amssymb}

\usepackage{mathtools, nccmath}
\DeclarePairedDelimiter{\nint}\lfloor\rceil

\usepackage{color}

\usepackage{multirow}
\usepackage{multicol}
\usepackage{booktabs}
\usepackage{adjustbox}
\usepackage{soul}

\newcommand\blfootnote[1]{%
  \begingroup
  \renewcommand\thefootnote{}\footnote{#1}%
  \addtocounter{footnote}{-1}%
  \endgroup
}

\makeatletter
\renewcommand*{\@fnsymbol}[1]{\ensuremath{\ifcase#1\or \dagger\or \ddagger\or
    \mathsection\or \mathparagraph\or \|\or **\or \dagger\dagger
    \or \ddagger\ddagger \else\@ctrerr\fi}}
\makeatother

\usepackage[pagebackref=true,breaklinks=true,letterpaper=true,colorlinks,bookmarks=false]{hyperref}

\cvprfinalcopy 

\def\cvprPaperID{20} 

\ifcvprfinal\fi
\begin{document}

\title{LSQ+: Improving low-bit quantization through learnable offsets and better initialization}

\author{Yash Bhalgat$^1$\\
\and
Jinwon Lee$^1$\\
\and
Markus Nagel$^2$\\
\and
Tijmen Blankevoort$^2$\\
\and
Nojun Kwak$^3$\thanks{Currently a Visiting Researcher at Qualcomm Technologies, Inc.}\\
\and
$^1$Qualcomm AI Research, Qualcomm Technologies, Inc.\\
$^2$Qualcomm AI Research, Qualcomm Technologies Netherlands B.V.\\
$^3$Seoul National University\\
{\tt\small \{ybhalgat, jinwonl, markusn, tijmen\}@qti.qualcomm.com,  nojunk@snu.ac.kr}\\
}


\maketitle
\thispagestyle{empty}

\begin{abstract}
  Unlike ReLU, newer activation functions (like Swish, H-swish, Mish) that are frequently employed in popular efficient architectures can also result in negative activation values, with skewed positive and negative ranges. Typical learnable quantization schemes \cite{pact2018, lsq} assume unsigned quantization for activations and quantize all negative activations to zero which leads to significant loss in performance. Naively using signed quantization to accommodate these negative values requires an extra sign bit which is expensive for low-bit (2-, 3-, 4-bit) quantization. To solve this problem, we propose LSQ+, a natural extension of LSQ \cite{lsq}, wherein we introduce a general asymmetric quantization scheme with trainable scale and offset parameters that can learn to accommodate the negative activations. Gradient-based learnable quantization schemes also commonly suffer from high instability or variance in the final training performance, hence requiring a great deal of hyper-parameter tuning to reach a satisfactory performance. LSQ+ alleviates this problem by using an MSE-based initialization scheme for the quantization parameters. We show that this initialization leads to significantly lower variance in final performance across multiple training runs. Overall, LSQ+ shows state-of-the-art results for EfficientNet and MixNet and also significantly outperforms LSQ for low-bit quantization of neural nets with Swish activations (e.g.: 1.8\% gain with W4A4 quantization and upto 5.6\% gain with W2A2 quantization of EfficientNet-B0 on ImageNet dataset). To the best of our knowledge, ours is the first work to quantize such architectures to extremely low bit-widths. 

\end{abstract}

\blfootnote{Qualcomm AI Research is an initiative of Qualcomm Technologies, Inc.}

\section{Introduction}
With the popularity of deep neural networks across various use-cases, there is now an increasing demand for methods that make deep networks run efficiently on resource-constrained edge-devices. These methods include model pruning, neural architecture search (NAS) and hand-crafted efficient networks made out of novel architectural blocks (e.g. depth-wise separable or group convolutions, squeeze-excite blocks, etc.). Finally we can also perform model quantization, where the weights and activations are quantized to lower bit-widths allowing efficient fixed-point inference and reduced memory bandwidth usage. 

Due to the surge in more efficient architectures found with NAS, newer and more general activation functions (like Swish \cite{swish}, H-swish \cite{mobilenetv3}, Leaky-ReLU) are replacing the traditional ReLU. Unlike ReLU, these activation functions also take over values below zero. Current state-of-the-art quantization schemes like PACT \cite{pact2018} and LSQ \cite{lsq} assume unsigned quantization ranges for activation quantization where all the activation values below zero are discarded by quantizing them to zero. This works well for traditional ReLU-based architectures like ResNet \cite{resnet2}, but leads to a significant loss of information when applied to modern architectures like EfficientNet \cite{efficientnet} and MixNet \cite{mixnet}, which employ Swish activations. For example, LSQ achieves W4A4 quantization of preactivation-ResNet50 with no loss in acccuracy but leads to a $4.1$\% loss in accuracy when quantizing EfficientNet-B0 to W4A4\footnote{W$x$A$x$ quantization indicates quantizing the weights and output activations of all layers to $x$ bits}. Naively using a signed quantization range to accommodate these negative values also results in a drop in performance. 

To alleviate these drops in performance which are commonly observed with very low-bit (2-, 3-, 4-bit) quantization, we propose using a general asymmetric quantization scheme with a learnable offset parameter as well as a learnable scale parameter. We show that the proposed quantization scheme learns to accommodate the negative activation values differently for different layers and recovers the accuracy loss incurred by LSQ, e.g. $1.8$\% accuracy improvement over LSQ with W4A4 quantization and upto $5.6$\% improvement with W2A2 quantization on EfficientNet-B0. To the best of our knowledge, ours is the first work to quantize modern architectures like EfficientNet and MixNet to extremely low bit-widths.

Another problem faced especially by any \textit{gradient-based} learnable quantization scheme is its sensitivity to initialization, meaning that a poor initialization can lead to a high variance in final performance across multiple training runs. This problem is especially observed with min-max initialization (used in \cite{tensorflow2015-whitepaper}). We show that using an initialization scheme based on mean-squared-error (MSE) minimization \cite{mse1,mse3} for the offset and scale parameters leads to significantly higher stability in final performance than min-max quantization. We also compare this initialization scheme with the one proposed in \cite{lsq}.

In summary, our proposed method, called LSQ+, extends LSQ \cite{lsq} by adding a simple yet effective learnable offset parameter for activation quantization to recover the lost accuracy on architectures employing \textit{Swish-like} activations. Furthermore, our other contribution is showing the importance of proper initialization for stable training, especially in the low-bit regime.




\section{Related Work}
A good overview of the basics of quantization is given in \cite{krishnamoorthi}, where the differences between asymmetric and symmetric quantization are explained. In general, we can classify quantization methods into \textit{post-training methods} that work without fine-tuning and \textit{quantization-aware training methods} that need fine-tuning. 

Post-training quantization methods \cite{banner2019,ocs2019,choukroun2019} optimize neural networks for quantization without full training and using a little amount of data. \cite{dfq,zeroq} do this better without using any data at all. Although these methods work well on typical 8-bit quantization, they were not able to achieve good accuracy on very low-bit (2, 3, 4-bit) quantization.

Quantization-aware training generally outperforms these methods on low-bit tasks given enough time to optimize.
Simulated quantization-aware training methods and improvements for these are discussed in \cite{Gupta2015,jacob2018cvpr,louizos2018relaxed}. Essentially, operations are added to the neural network computational graph that simulate how quantization would be done on an actual device. Several recent papers improve over these methods by learning the quantization parameters, e.g. QIL \cite{qil}, TQT \cite{tqt} and LSQ \cite{lsq}. This is the approach we build upon in our paper, but a similar asymmetric quantization scheme and initialization we suggest could be used for any other methods. 

In a parallel line of research, some works \cite{qkd,mishra2017apprentice,polino2018model} have tried to apply knowledge distillation to quantization resulting in improved performances. Also, some recent work \cite{dq2020} has been done on automatically learning the bit-width alongside of the ranges. Note that our proposed method is orthogonal to these works, and thus it can be jointly used with them. Lastly, several papers have introduced different quantization grids than uniform one we use. In \cite{miyashita2016convlog} and \cite{ullrich2018}, a logarithmic space or fully free-format quantization space are used to quantize the network. In this paper, we do not consider this, as the hardware implementations for these are simply inefficient, requiring costly lookup table or approximation on runtime.

\section{Method}
\label{sec:method}
In LSQ \cite{lsq}, a symmetric quantization scheme with a trainable scale parameter is proposed for both weights and activations. This scheme is defined as follows:
\begin{align}
\begin{split}
    \bar{x} &= \nint*{clamp\left(\frac{x}{s}, n, p\right)} \\
    \hat{x} &= \bar{x}\times s
\end{split}
\label{eq:symm}
\end{align}
where $\nint{\cdot}$ indicates the round function and the $clamp(\cdot)$ function clamps all values between $n$ and $p$. $\bar{x}$ and $\hat{x}$ denote the coded bits and quantized values, respectively. 
LSQ can make use of a signed or an unsigned quantization range. However, both are suboptimal for activation functions like Swish or Leaky-ReLU which have skewed negative and positive ranges\footnote{For example, the negative portion of Swish activation lies only between $-0.278$ and $0$ whereas the positive portion is unbounded.}. 
Using an unsigned quantization range, i.e. $n=0, p=2^{b}-1$, clamps all negative activations to zero leading to a significant loss of information. 
On the contrary, using a signed quantization range, i.e. $n=-2^{b-1}, p=2^{b-1}-1$, will quantize all negative activations to integers in the range $[-2^{b-1},0]$ and all positive activations to $[0,2^{b-1}-1]$, hence giving equal importance to the negative and positive portions of the activation function. 
However, this loses valuable precision for skewed distributions where the positive dynamic range is significantly larger than the negative one. In Sec. \ref{results:swish}, we will show that both quantization schemes lead to a significant loss in accuracy when quantizing architectures with Swish activations.

The proposed method LSQ+ solves the above mentioned problem with a more general learnable asymmetric quantization scheme for the activations, described in Sec. \ref{sec:asymm_formulation}. Sec. \ref{sec:init} describes the initialization scheme used in LSQ+. 


\subsection{Learnable asymmetric quantization}
\label{sec:asymm_formulation}
As a solution to the above mentioned problem, we propose a general asymmetric activation quantization scheme where not only the scale parameters but also the offset parameters are learned during training to handle skewed activation distributions:
\begin{align}
\begin{split}
    \bar{x} &= \nint*{clamp\left(\frac{x-\beta}{s}, n, p\right)} \\
    \hat{x} &= \bar{x}\times s + \beta
\end{split}
\label{asymm}
\end{align}

Here, the offset parameter $\beta$ and the scale $s$ are both learnable.
The gradient update of the parameter $s$ is calculated using:
\begin{align}
\begin{split}
    \frac{\partial \hat{x}}{\partial s} &=\frac{\partial \bar{x}}{\partial s} s+\bar{x} \\&\simeq 
    \left\{\begin{array}{ll}
        -\displaystyle\frac{x-\beta}{s}+\displaystyle\nint*{\frac{x-\beta}{s}} & \text { if } n<\displaystyle\frac{x-\beta}{s}<p \\
        n \text{ or } p & \text { otherwise. }
    \end{array}\right.
\end{split}
\label{eq:ds}
\end{align}
And the gradient update of $\beta$ is calculated using:
\begin{align}
\frac{\partial \hat{x}}{\partial \beta}=\frac{\partial \bar{x}}{\partial \beta} s+1 \simeq \left\{\begin{array}{ll}
0 & \text { if } n<(x-\beta) / s<p \\
1 & \text { otherwise. }
\end{array}\right.
\label{eq:dbeta}
\end{align}
In both (\ref{eq:ds}) and (\ref{eq:dbeta}), straight-through-estimator (STE) \cite{straight_through} is used in approximating $\partial \bar{x} / \partial s$ and $\partial \bar{x} / \partial \beta$.

For weight quantization, we use symmetric signed quantization (\ref{eq:symm}) since the layer weights can be empirically observed to be distributed symmetrically around zero. Because of this, asymmetric quantization of activations has no additional cost during inference as compared to symmetric quantization since the additional offset term can be precomputed and incorporated into the bias at compilation time:
\begin{align}
\label{eq:overhead}
\hat{w}\hat{x}=(\bar{w}\times s_w)(\bar{x}\times s_x+\beta)=\bar{w}\bar{x}s_w s_x+\underbrace{\beta s_w \bar{w}}_{bias}.    
\end{align}

\begin{table*}[t]
	\caption{Different possible parametrizations for LSQ+'s learnable asymmetric quantization scheme}
	\centering
	\begin{tabular}{l | c| c | c | c}
        \toprule
        	{Configuration} & {$s$} & {$\beta$} & {$n$} & {$p$} \\
			
			\midrule
	
			Config 1 : Unsigned + Symmetric (LSQ) & trainable & N/A & 0 & $2^{b}-1$\\
			Config 2 : Signed + Symmetric & trainable & N/A & $-2^{b-1}$ & $2^{b-1}-1$  \\
			Config 3 : Signed + Asymmetric & trainable & trainable & $-2^{b-1}$ & $2^{b-1}-1$ \\
			Config 4 : Unsigned + Asymmetric & trainable & trainable & 0 & $2^{b}-1$ \\
        \bottomrule
	\end{tabular}
	\label{configurations}
\end{table*}

Table \ref{configurations} shows four possible parametrizations for the proposed quantization scheme in (\ref{asymm}). Configurations 1 and 2 do not use an offset parameter, hence following the learnable symmetric quantization scheme proposed in LSQ \cite{lsq}. Since Configuration 1 uses an unsigned range with this symmetric quantization scheme, it corresponds exactly to the parametrization proposed in LSQ for activation quantization. 
Configurations 3 and 4 learn both the scale and offset parameter for activation quantization, the only difference being signed and unsigned quantization ranges. We will analyze these different parametrizations in the experiments section.

\subsection{Initialization of quantization parameters}
\label{sec:init}
As we enter the \textit{extremely low bit-width} regime with gradient-based learnable quantization methods, the final performance after training becomes highly sensitive to the initialization of the quantization hyperparameters. This sensitivity problem is amplified in the presence of depthwise separable convolutions which are known to be challenging to quantize \cite{haq}. In this work, we propose an initialization scheme for the scale and offset parameters that achieves significantly more stable and sometimes better performance than other initializations (while keeping the quantization configuration unchanged) proposed in the literature \cite{jacob2018cvpr, lsq}.

\subsubsection{Scale initialization for weight quantization} 
\begin{figure}
    \centering
    \includegraphics[width=0.7\linewidth, height=0.5\linewidth]{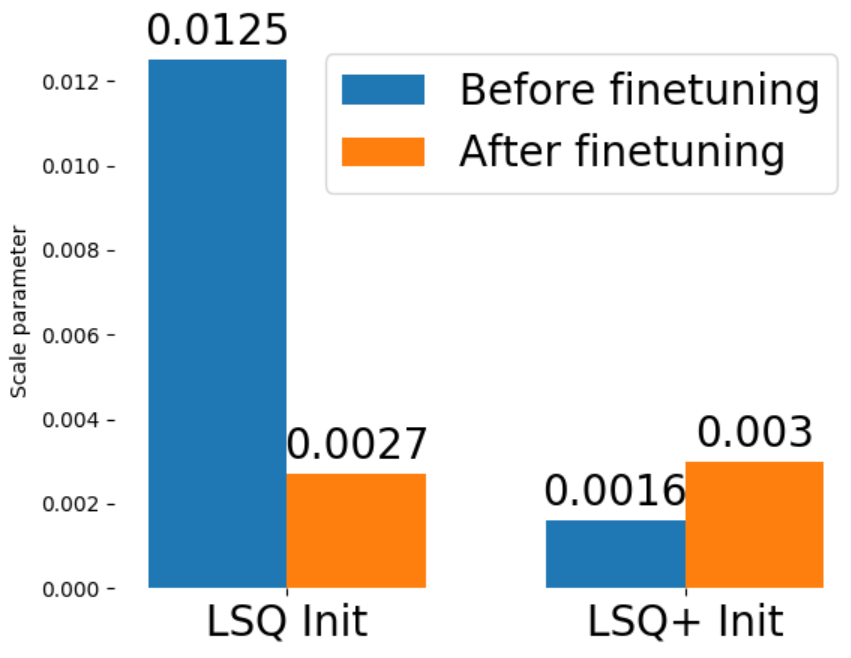}
    \caption{Figure shows the scale parameter of weight quantizer in \textbf{blocks.1.conv.0} layer of EfficientNet-B0 before and after finetuning with LSQ and LSQ+ initializations. For both experiments, we used configuration 4 for activation quantization. As shown, LSQ init of the scale is further from the converged value as compared to LSQ+. More on effects of initialization in Sec \ref{results:init}}
    \label{fig:scale_init}
\end{figure}

As mentioned before, we use signed symmetric quantization for the weights (similar to Configuration 2) in our method. Hence, no offset is used for weight quantization. LSQ \cite{lsq} proposes using the square-root normalized average absolute value of layer weights, i.e. $2\langle |w| \rangle/\sqrt{p}$, to intialize the scale parameter. This leads to a very large initialization for 2-, 3- or 4-bit quantizaiton, e.g. $s_{init}=\langle |w|\rangle/\sqrt{2}$ for 4-bit case. From our experiments, this initialization was observed to be far from the converged values of the scale parameters. One of the instances of this phenomenon is shown in Figure \ref{fig:scale_init}.

We fix this problem by using the statistics of the weight distribution rather than the actual weight values for the initialization. Similar to \cite{dfq}, we use a Gaussian approximation for the weight distribution in each layer. Following this, we initialize the scale parameter for each layer by:
$$s_{init} = max(|\mu-3*\sigma|, |\mu+3*\sigma|)/2^{b-1}$$
where $\mu$ and $\sigma$ are the mean (same as $\langle|w|\rangle$) and standard deviation of the weights in that layer.

\subsubsection{Scale/offset initialization for activation quantization}
Let $x_{min}$ and $x_{max}$ denote the min and the max value of the activation function. For example, $x_{min}=0$ for ReLU and $x_{min}=-0.278$ in case of Swish activations\footnote{For unbounded activation functions (e.g. positive portion of Swish), $x_{min}$ or $x_{max}$ can be estimated from a few forward passes.}. Intuitively, a full utilization of the quantization range can be obtained when $x_{min}$ is quantized to the lower bound of the quantization range and $x_{max}$ to the upper bound. Following this intuition, an initialization for $s$ and $\beta$ would satisfy:
\begin{align}
    \frac{x_{min}-\beta_{init}}{s_{init}}\xrightarrow{}n \quad , \quad \frac{x_{max}-\beta_{init}}{s_{init}}\xrightarrow{}p.
    \label{minmax}
\end{align}
Solving these constraints yields:
\begin{align}
    s_{init}=\frac{x_{max}-x_{min}}{p-n} \quad ,  \beta_{init}=x_{min}-n*s_{init}.
    \label{minmaxinit}
\end{align}

But the above initialization is highly prone to outliers in the activation distribution, especially since the activation ranges are dynamic. To overcome this, we propose initializing the scale and offset parameters per layer by optimizing the MSE minimization problem, similar to \cite{mse1,mse3}:
\begin{align}
    s_{init}, \beta_{init} = \underset{s,\beta}{\arg\min} ||\hat{x}-x||^2_F
    \label{mseinit}
\end{align}where $\hat{x}$ is given by (\ref{asymm}). There is no closed-form solution to (\ref{mseinit}). Hence, we embed equations (\ref{eq:ds}) and (\ref{eq:dbeta}) into PyTorch's autograd functionality to optimize for $\{s_{init},\beta_{init}\}$ over a few batches of data.

\begin{table*}[t]
	\caption{Comparison of all configurations of quantization with EfficientNet-B0 (FP accuracy: 76.1\%)}
	\centering
	\begin{tabular}{l | c| c | c }
        \toprule
        	{Method} & {W2A2} & {W3A3} & {W4A4} \\
			
			\midrule
	
			Config 1 : LSQ (Unsigned + Symmetric) & 43.5\% & 67.5\% & 71.9\% \\
			Config 2 : Signed + Symmetric & 23.7\% & 54.8\% & 68.8\%  \\
			Config 3 : Signed + Asymmetric & \textbf{49.1\%} & \textbf{69.9\%} & 73.5\%  \\
			Config 4 : Unsigned + Asymmetric & 48.7\% & 69.3\% & \textbf{73.8\%} \\
        \bottomrule
	\end{tabular}
	\label{table:efficientnet}
\end{table*}

\begin{table*}[t]
	\caption{Comparison of all configurations of quantization with MixNet-S (FP accuracy: 75.9\%)}
	\centering
	\begin{tabular}{l | c| c | c }
        \toprule
        	{Method} & {W2A2} & {W3A3} & {W4A4} \\
			
			\midrule
	
			Config 1 : LSQ (Unsigned + Symmetric) & 39.9\% & 64.3\% & 70.4\% \\
			Config 2 : Signed + Symmetric & 23.4\% & 62.1\% & 67.2\%  \\
			Config 3 : Signed + Asymmetric & 42.5\% & \textbf{66.7\%} & 71.6\%  \\
			Config 4 : Unsigned + Asymmetric & \textbf{42.8\%} & 66.1\% & \textbf{71.7\%} \\
        \bottomrule
	\end{tabular}
	\label{table:mixnet}
\end{table*}

\begin{table*}[t]
	\caption{Comparison of all configurations of quantization with ResNet18 (FP accuracy: 70.1\%)}
	\centering
	\begin{tabular}{l | c| c | c }
        \toprule
        	{Method} & {W2A2} & {W3A3} & {W4A4} \\
			
			\midrule
	
			PACT \cite{pact2018} & 64.4\% & 68.1\% & 69.2\% \\
			DSQ \cite{dsq} & 65.2\% & 68.7\% & 69.6\% \\
			QIL \cite{qil} & 65.7\% & 69.2\% & 70.1\% \\
			Config 1 : LSQ (Unsigned + Symmetric) & 66.7\% & \textbf{69.4\%} & 70.7\% \\
			Config 2 : Signed + Symmetric & 64.7\% & 66.1\% & 69.2\%  \\
			Config 3 : Signed + Asymmetric & 66.7\% & \textbf{69.4\%} & 70.7\%  \\
			Config 4 : Unsigned + Asymmetric & \textbf{66.8\%} & 69.3\% & \textbf{70.8\%} \\
        \bottomrule
	\end{tabular}
	\label{table:resnet}
\end{table*}

\section{Experiments}
We evaluate the effectiveness of our method by quantizing architectures with Swish activations to W2A2, W3A3 and W4A4. To the best of our knowledge, ours is the first work to quantize such architectures to extremely low bit-widths. As a sanity check, we show that LSQ+ also maintains the performance of LSQ \cite{lsq} on traditional architectures with ReLU activation function. Finally, we show the effect of using different initializations on the performance of the proposed quantization method. All experiments are performed on the ImageNet \cite{ILSVRC15} dataset.

In all configurations and all experiments, the weight parameters are initialized with the pretrained floating point weights of the deep network. Although we will compare the effectiveness of different initializations for the scale/offset parameters in Sec \ref{results:init}, we use our proposed initialization from Sec \ref{sec:init}  for experiments in sections \ref{results:swish} and \ref{results:relu}.

\subsection{Results on Swish activation}
\label{results:swish}
Tables \ref{table:efficientnet} and \ref{table:mixnet} show the performance impact of quantization with all the configurations of the proposed method on EfficientNet-B0 \cite{efficientnet} and MixNet-S \cite{mixnet}, respectively. MixNet-S uses ReLU activation in the initial 3 layers and Swish activation in rest of the layers. By using the learnable offset parameter, we observe a $1.6$-$1.8$\% and $1.2$-$1.3$\% performance improvement for W4A4 quantization on EfficientNet-B0 and MixNet-S respectively (see Configurations 3 and 4 compared to Configuration 1 (LSQ)). This performance improvement using our proposed learnable asymmetric quantization scheme is most prominent in the case of W2A2 quantization.

The performance of Configuration 3 (signed range + learnable offset) and 4 (unsigned range + learnable offset) is almost similar for all the bit-widths. This is because, since we learn the offset parameter, the activation range is appropriately mapped to the quantization range irrespective of it being signed or unsigned.

Another interesting observation is that Configuration 2 performs consistently worse than all other configurations. This is because, due to the lack of an offset parameter, only $2^{b-1}$ quantization levels are utilized by the positive part of the activation range while the positive portion of the Swish activation is much larger than the negative portion, as mentioned in Section \ref{sec:method}. Hence, compared to Configurations 3 and 4 which allocate $2^b$ quantization levels for the entire activation range, Configuration 2 has a poor utilization of its quantization range, leading to a worse performance.

\subsection{Results on ReLU activation}
\label{results:relu}

The results on ResNet shown in the LSQ paper \cite{lsq} use the pre-activation version of ResNet architecture \cite{resnet2} which has about $0.4$-$0.6$\% higher top-1 ImageNet accuracy than the standard ResNet(s). Hence, for a fair comparison with other state-of-the-art methods, we run our own implementation of LSQ (Configuration 1) and all other configurations on the standard ResNets. Tables \ref{table:resnet} shows the quantization performance of all the configurations of the proposed method on ResNet18. Our implementation of LSQ (Configuration 1) can achieve a $70.7$\% accuracy with W4A4 quantization which is more than full-precision accuracy of $70.1$\%. This is sanity check that proves that our LSQ results are \textit{at par} with the original LSQ paper \cite{lsq}. Also, Configurations 1, 3 and 4 outperform existing state-of-the-art methods, namely PACT \cite{pact2018}, DSQ \cite{dsq} and QIL \cite{qil}. It is worth noting that, unlike EfficientNet and MixNet, there is almost no performance gap between Configurations 1, 3 and 4 when quantizing ResNet18. We attribute this to the fact that ReLU activation function has no negative component.

\subsection{Effect of quantization parameter initialization}
\label{results:init}
In this section, we compare three schemes for initializing the quantization scale and offset parameters. Since we use symmetric quantization for weights, no offset is used for weight quantization. Also, configurations 1 and 2 for activation quantization don't use an offset.
The three compared initialization methods are as follows:

\begin{enumerate}
\itemsep0pt
    \item \textbf{Min-max initialization.} We use the minimum and maximum values of each layer's weights and activations (obtained over first batch of input images) to initialize the quantization scale and offset parameters. This initialization scheme is formalized in (\ref{minmaxinit}).
    \item \textbf{LSQ initialization.} The scale for both weight quantization and activation quantization is initialized as $2*mean(|v|)/\sqrt{p}$, where $v$ indicate layer weights or activations and $p$ is the upper bound of the quantization range.
    \item \textbf{LSQ+ initialization.} We intialize the weight quantization and activation quantization parameters as proposed in Sec. \ref{sec:init}
\end{enumerate}

For the experiments, we quantize EfficientNet-B0 using Configuration 4 and perform multiple training runs with each of these initialization methods. Table \ref{table:sensitivity} shows the variation ($\Delta_{acc}$) in the final performance across 5 training runs with each of these initializations. We can observe a high instability in the final performance with W2A2 quantization, especially with min-max quantization. This is because the tail of the weight or activation distribution can easily influence the scale parameter intialization with 2-bit quantization. The LSQ initialization method, which initializes the weight quantization scale parameter with the square-root normalized mean absolute value, also has a higher variation in training performance. This is because LSQ initialization leads to a large value for the $s_{init}$ which is far from the converged value as was shown in Figure \ref{fig:scale_init}.


\begin{table}[t]
	\caption{$\Delta_{acc}$ around mean accuracy across 5 training runs for EfficientNet quantization using Config 4 with different initializations. Note: other tables show the \textit{best} accuracy after grid search on hyperparameters, which is different from mean accuracy.}
	\centering
	\begin{tabular}{c| c | c}
        \toprule
        	\multirow{2}{*}{\shortstack{Quantization Parameter\\Initialization}} & \multicolumn{2}{c}{Mean Acc $\pm\Delta_{acc}$}\\
        	\cmidrule{2-3}
        	& {W4A4} & {W2A2} \\
			
			\midrule
	
			Mix-max & $71.3\pm2.2$\% & $43.8\pm4.7$\% \\
			LSQ & $72.0\pm1.6$\% & $44.4\pm2.9$\% \\
			LSQ+ & $73.0\pm\textbf{0.9}$\% & $46.8\pm\textbf{1.9}$\% \\
        \bottomrule
	\end{tabular}
	\label{table:sensitivity}
\end{table}

			
	


\section{Discussion}
\label{discussion}


\subsection{Learned offset values}
\begin{figure}[h]
    \centering
    \includegraphics[width=\linewidth]{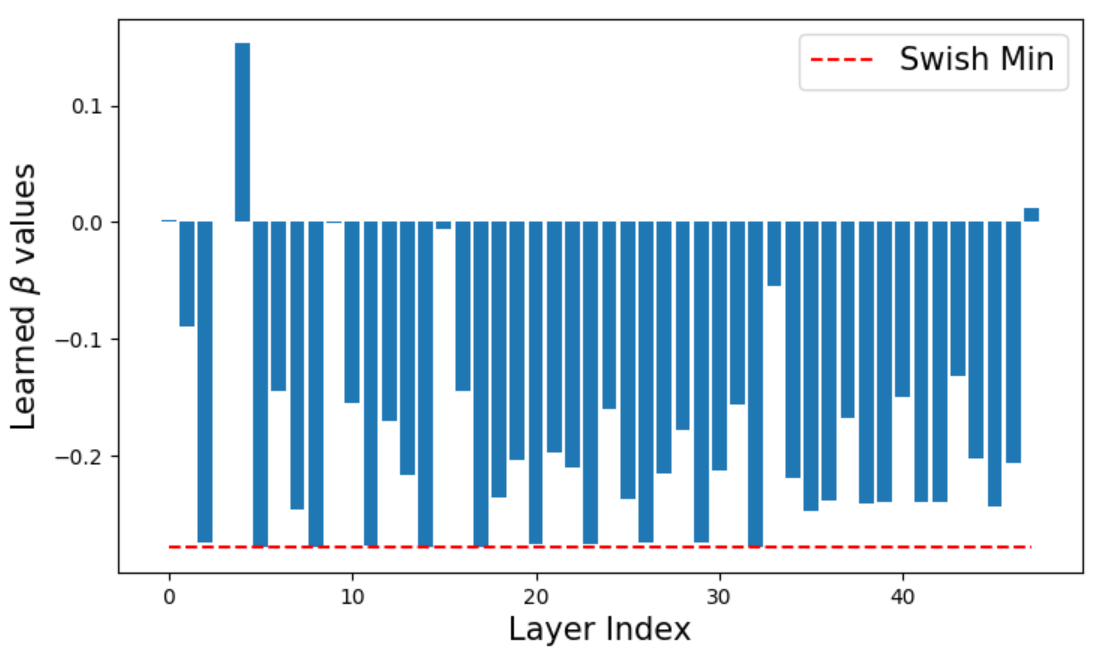}
    \caption{Layerwise $\beta$ values after covergence for EfficientNet-B0 }
    \label{fig:layerwise_beta}
\end{figure}

It is interesting to observe the layer-wise offset values learned by the network. 
Figure \ref{fig:layerwise_beta} shows one such example with Configuration 4 for W4A4 quantization of EfficientNet-B0. Note that an offset is not used for quantizing the squeeze-excite layers because \textit{sigmoid} activation function has no negative component. Also, there is no activation applied at the end of a bottleneck block in EfficientNet, hence we use symmetric-signed-quantization for those activation layers. These layers are not shown in the plot.

We can observe that most of the $\beta$ values are negative, meaning that the activations are shifted ``up" before being scaled and clamped between the quantization range. This shows that the quantization layers learn to accommodate the negative activation values. None of the learned $\beta$ values are lower than the min value of the Swish activation function (red dotted line). Because, from (\ref{eq:dbeta}), $$\beta<x_{min}\implies\frac{x-\beta}{s}>0 \;\,\forall x>x_{min} \implies \frac{\partial \hat{x}}{\partial \beta} =0$$
Hence, gradient for $\beta$ becomes zero as soon as $\beta<x_{min}$.

\subsection{Learned vs Fixed offset}
On further observation of Figure \ref{fig:layerwise_beta}, the learned offset for most layers is away from the Swish minimum value. This is because, if we try to represent the entire activation range using the quantization grid (refer (\ref{minmax})), it leads to coarser representation since the number of bits are fixed causing a higher quantization error. The purpose of learning the $s$ and $\beta$ values is to learn this trade-off between resolution of the quantization grid and the proportion of activation range represented by the quantization grid. Hence, the learned $\beta$ values are not exactly equal to the min value of the activation function. But one might wonder about the performance achieved when $\beta$ for each layer is fixed to $x_{min}$. Table \ref{table:fixed_offset} shows the difference of performance between fixed and learned offset methods.
\begin{table}[t]
	\caption{Performance difference between fixed and learned offset for EfficientNet quantization at W4A4 using Config 4}
	\centering
	\begin{tabular}{c| c }
        \toprule
        	Method & {W4A4} \\
			
			\midrule
			Fixed $\beta=0$ (LSQ) & $71.9$\% \\
			Fixed $\beta=x_{min}$ & $72.5$\% \\
			Learned $\beta$ & $73.8$\% \\
        \bottomrule
	\end{tabular}
	\label{table:fixed_offset}
\end{table}


\section{Conclusion}
In this work, targeting the low-bit quantization domain, we solve two problems: (1) quantization of deep neural networks with signed activation functions and (2) stability of training performance w.r.t. quantization. To do so, we propose a general asymmetric quantization scheme with trainable scale and offset parameters that can learn to accommodate the negative activations without using an extra sign bit. In (\ref{eq:overhead}), we show that using such asymmetric quantization for activations incurs zero runtime overhead. Our work is the first to quantize modern efficient architectures like EfficientNet and MixNet to extremely low bits. We show that LSQ+ significantly improves the performance of 2-, 3- and 4-bit quantization on these architectures. Our experiments with traditional ReLU-based ResNet18 architecture show that we can use LSQ+ instead of LSQ everywhere without hurting performance. Finally, we show that using MSE-minimization based initialization scheme for the activation quantization parameters leads to a more stable performance, which is of high importance for low-bit quantization-aware training.

{\small
\bibliographystyle{ieee_fullname}
\bibliography{egbib}
}

\end{document}